\documentclass{ieeeaccess}
\usepackage{cite}
\usepackage{amsmath,amssymb,amsfonts}
\usepackage{graphicx}
\usepackage{textcomp}
\def\BibTeX{{\rm B\kern-.05em{\sc i\kern-.025em b}\kern-.08em
		T\kern-.1667em\lower.7ex\hbox{E}\kern-.125emX}}
	
\usepackage{float}
\usepackage{booktabs}
\usepackage[dvipsnames]{xcolor}
\usepackage[normalem]{ulem}
\usepackage{color,soul}
\usepackage{adjustbox}
\usepackage[lofdepth,lotdepth]{subfig}
\usepackage{algorithm}
\usepackage{algpseudocode}
\usepackage{multirow}
\algrenewcommand\algorithmicrequire{\textbf{Input}}
\algrenewcommand\algorithmicensure{\textbf{Output}}
\renewcommand{\thetable}{\Roman{table}}

\usepackage{tabularx}
\newcolumntype{b}{>{\hsize=1.2\hsize}X}
\newcolumntype{s}{>{\hsize=.8\hsize}X}	

\begin{document}
\history{Date of publication xxxx 00, 0000, date of current version xxxx 00, 2022.}
\doi{10.1109/ACCESS.2017.DOI}

\title{Clustering and Analysis of GPS Trajectory Data using Distance-based Features}
\author{\uppercase{Zann Koh}\authorrefmark{1}, 
	\uppercase{Yuren Zhou}\authorrefmark{1}, \IEEEmembership{Member, IEEE}, \uppercase{Billy Pik Lik Lau}\authorrefmark{1}, \IEEEmembership{Member, IEEE} ,\uppercase{Ran Liu}\authorrefmark{1}, \uppercase{Keng Hua Chong} \authorrefmark{2}, and \uppercase{Chau Yuen}\authorrefmark{1}, \IEEEmembership{Fellow, IEEE}
\address[1]{Engineering and Product Development, Singapore University of Technology and Design (SUTD), Singapore 487372}
\address[2]{Architecture and Sustainable Design, Singapore University of Technology and Design (SUTD), Singapore 487372}}
\tfootnote{This research is supported by the Singapore Ministry of National Development and the National Research Foundation, Prime Minister’s Office under the Land and Liveability National Innovation Challenge (L2 NIC) Research	Programme (L2 NIC Award No. L2NICTDF1-2017-4). Any opinion, findings, and conclusions or recommendations expressed in this material are those of the author(s) and do not reflect the views of the Singapore Ministry of National Development and National Research Foundation, Prime Minister’s Office, Singapore.}

\markboth
{Koh \headeretal: Clustering and Analysis of GPS Trajectory Data using Distance-based Features}
{Koh \headeretal: Clustering and Analysis of GPS Trajectory Data using Distance-based Features}

\corresp{Corresponding author: Zann Koh (e-mail: zann\_koh@mymail.sutd.edu.sg).}
	
\begin{abstract}
The proliferation of smartphones has accelerated mobility studies by largely increasing the type and volume of mobility data available. One such source of mobility data is from GPS technology, which is becoming increasingly common and helps the research community understand mobility patterns of people. However, there lacks a standardized framework for studying the different mobility patterns created by the non-Work, non-Home locations of Working and Nonworking users on Workdays and Offdays using machine learning methods. We propose a new mobility metric, Daily Characteristic Distance, and use it to generate features for each user together with Origin-Destination matrix features. We then use those features with an unsupervised machine learning method, $k$-means clustering, and obtain three clusters of users for each type of day (Workday and Offday). Finally, we propose two new metrics for the analysis of the clustering results, namely User Commonality and Average Frequency. By using the proposed metrics, interesting user behaviors can be discerned and it helps us to better understand the mobility patterns of the users.
\end{abstract}
	
	
\titlepgskip=-15pt
	
\maketitle
	
\section{Introduction}
\label{sec:introduction}
\PARstart{G}{lobal} Positioning System, or GPS for short, has been around for many years and is increasingly being used in the context of mobility studies. It has been found to be widely usable for collection of spatio-temporal data on different scales \cite{van2009sensing}. GPS mobility data has been used in many different fields and applications, such as finding efficient routes \cite{ta2016built}, understanding the progression of infectious diseases\cite{hast2019use}, and prediction or inferring of demographic information of users \cite{solomon2018predict,wu2019inferring}. 

Many studies analyze GPS data in conjunction with other data, such as demographic data\cite{sila2016analysis,long2021daily}, supplementary survey data \cite{van2009sensing}, Wi-Fi and geolocation data\cite{brouwers2013dwelling}, or even sound and light data \cite{marakkalage2018understanding}. With increasing privacy concerns in recent years, it has become more difficult to obtain such data for large numbers of volunteers. Additionally, large volumes of human movement data are created without such supplementary data. To be eventually able to make use of this, we want to explore ways in which we can analyze GPS data without the need for additional supplementary data. Zhu \textit{et al.} \cite{zhu2017prediction} found that the user's socio-demographic role can be predicted with high accuracy using long-term GPS data, which supports the idea that Working and Nonworking users may have different mobility patterns. In addition to this, Nahmias-Biran \textit{et al.} \cite{nahmias2018enriching} found several distinct clusters of activity-travel patterns in their GPS-enriched travel survey dataset, which included distinct temporal patterns of different Out-of-Work activities as well as different Leisure activities. This leads us to examine the mobility patterns of Workdays and Offdays separately.

Although there are many works that have used GPS data in many different applications, there lacks a study that compares the mobility patterns of Working and Nonworking users, with a focus on non-Home, non-Work locations, on Workdays as compared to Offdays. Some challenges faced in this research are ensuring a fair comparison between users who live and work at different locations, as well as a fair comparison between Workdays and Offdays. To this end, we propose a new mobility metric that excludes the effects of Home and Work locations and uses the user's Home location as a reference point. We decided to use an unsupervised machine learning method - clustering, which finds groups in data without the need for labels or ground truth. We use the above-mentioned metric for each user in conjunction with other features as an input for the clustering algorithm. 

Therefore, this paper has the following contributions:
\begin{itemize}
	\item We propose a new mobility metric, Daily Characteristic Distance (DCD), as a fair basis to compare the mobility of Working and Nonworking users on Workdays and Offdays separately, even with differing distances between the Home and Work locations for different users.
	\item We use the DCD to extract features from users and use these features in conjunction with Origin-Destination (OD) matrix features to cluster users using $k$-means clustering on a real-world dataset collected in Singapore.
	\item Finally, we analyze the cluster results using two other analysis metrics that we have proposed - User Commonality and Average Frequency, which utilize information from within the clusters to gather higher-level insights.
	
\end{itemize}

The structure of the remaining sections will be as follows: Section II lists some related works in the field of GPS tracking and clustering. The dataset and preprocessing procedures are presented in Section III, while the proposed methodology is presented in Section IV. Section V shows the results and analysis of performing our proposed methodology on Workday data, while Section VI does the same for Offday data. Lastly, Section VII concludes the paper.

\section{Related Works}
This section will be split into three parts. The first part addresses past works on the analysis of human mobility via the usage of GPS data. The second part addresses the selection of clustering algorithms used for this paper. Lastly, the third part deals with mobility metrics.

\begin{figure*}[bt]
	\centering
	\includegraphics[width=\linewidth]{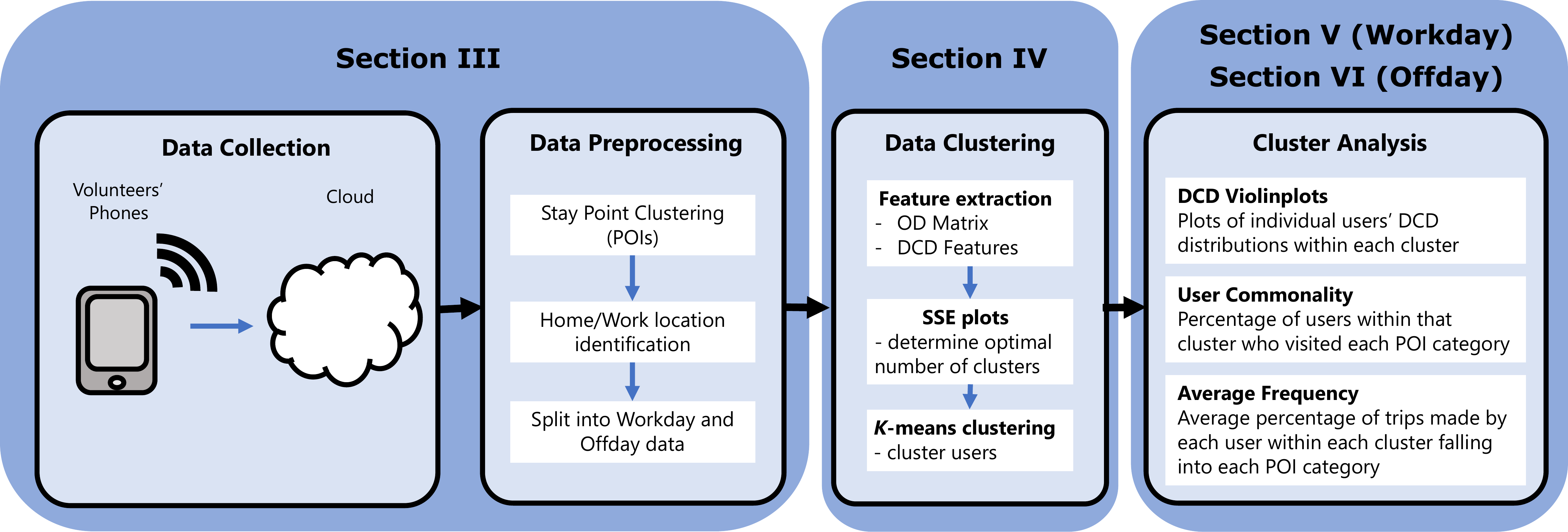}
	\caption{\textit{Flowchart depicting the data collection, processing, clustering, and analysis framework proposed by this paper.}}
	\label{fig:flowchart}
\end{figure*}

\subsection{Works Using GPS Data}
There have been several works focusing on the use of GPS data in mobility studies. Van der Spek \textit{et al.} \cite{van2009sensing} used GPS to collect data in three European city centers, as well as track the activity data of 13 families in Almere (a city in The Netherlands) for one week and conclude that GPS offers wide usability in the collection of invaluable spatial-temporal data on different scales and in different settings, adding new layers of knowledge to urban studies. 

Some studies make use of external data to supplement GPS data in order to gain additional insights. Sila-Nowicka \textit{et al.} \cite{sila2016analysis} performed an analysis of significant places identified from their GPS data in conjunction with additional social demographic data, while Long and Reuschke \cite{long2021daily} even use detailed GPS data to analyze the effects of employment type (business owners or employees) on daily mobility. Marakkalage \textit{et al.} \cite{marakkalage2021wifi} used a fusion of GPS data and Wi-Fi data to derive insights on neighbourhood activity and micro mobility.
 
GPS datasets may also provide an avenue for modelling human mobility if they are large enough. Alessandretti \textit{et al.} \cite{alessandretti2017multi} presented an extensive characterization of the statistical properties of GPS trajectories using a dataset collected from around 850 people lasting around 25 months, while Solmaz \textit{et al.} \cite{solmaz2017modeling} used GPS traces to model and simulate pedestrian mobility in disaster areas. 
 
Other machine learning approaches in the analysis of GPS data include supervised learning, where Zheng \textit{et al.} \cite{zheng2008understanding} proposed an approach based on supervised learning to infer people's motion modes from their GPS logs, as well as anomaly detection, where Wang \textit{et al.} \cite{wang2018detecting} proposed a hierarchical clustering method using an improved edit distance algorithm to detect anomalous taxi trajectories between selected pairs of origins and destinations. 

From the above, we understand that GPS data can be a rich source of mobility information about users, even extending to other aspects of demographics. As our focus is more on unsupervised machine learning as compared to prediction, we turn our focus to the application of clustering methods in the analysis of GPS data. 

\subsection{Clustering Algorithms in the Analysis of GPS Data}
Some authors have performed clustering on trajectories to find common routes or popular locations. An example would be Cesario \textit{et al.} \cite{cesario2017approach}, who proposed their own algorithm, Trajectory Pattern Mining (TPM), and used it to discover dense regions and popular sequential patterns within their dataset. Kumar \textit{et al.} \cite{kumar2018fast} also proposed their own novel algorithm, with the aim of discovering clusters of taxi routes. Dodge \textit{et al.} \cite{dodge2012movement} proposed their own framework to assess movement similarity using symbolic representation of movement parameters and used it to cluster trajectories of hurricanes and couriers according to their proposed similarity metrics. Tang \textit{et al.} \cite{tang2015uncovering} used the Density-Based Spatial Clustering of Applications with Noise (DBSCAN) \cite{ester1996density} algorithm to cluster locations of pick up and drop off points for their taxi GPS dataset, aiming to describe the taxi trips using statistical models. These above clustering methods are applied to the trajectories themselves and not to the users, which is less aligned with the aims of our study of the users' mobility patterns on Working and Nonworking days.

Another form of clustering in the analysis of GPS data is the grouping of users based on their travel patterns.  Amichi \textit{et al.} \cite{amichi2019mobility} used Gaussian mixture models \cite{reynolds2009gaussian} to identify three different groups of people based on their travel patterns - scouters, regulars, and routineers. However, this was mostly based on how often each user traveled to new locations as compared to returning to previously visited ones. In the long term, the number of recorded ``visited" places will keep increasing, while the ``new" locations will become few and far between, so this approach may not be applicable on a long term scale and is less suited for this study.

Scherrer \textit{et al.} \cite{scherrer2018travelers} went through a rigorous selection process for parameters and algorithms before performing their clustering. Out of a total of four clustering algorithms, $k$-means clustering\cite{lloyd1982least} was in at least the first two combinations in terms of their overall ranking, and they eventually used it to cluster users based on the large amount of data they gathered from a mobile application. They were able to draw conclusions from data that was gathered as a byproduct and hence without specific experimental aim or ground truth. As this is similar to our use case, we eventually decided to use $k$-means clustering.

\subsection{Existing Mobility Metrics}
We aim to find some simple, understandable features within our data that will result in meaningful interpretation of the clustering results. The review paper by Solmaz \textit{et al.} \cite{solmaz2019survey} classifies mobility metrics into three different types - movement-based, link-based, and network-based metrics. As our focus is on how the users travel, we place an emphasis on movement-based and link-based metrics such as visit frequency and mean squared distance, rather than on the network-based metrics such as transmission count and energy consumption.

Movement-based metrics include visit frequency and mean squared distance. A commonly used metric that combines these is radius of gyration \cite{gonzalez2008understanding}, which has been used in many papers \cite{pappalardo2015returners,pepe2020covid,xu2018human}. It gives the characteristic distance traveled by a user within a specified time period and is calculated as the mean squared distance of the user's visited locations to the center of mass of those locations. We are interested in non-Home and non-Work locations, thus we adapted this formula based on what we have in our dataset to extract the relevant features for clustering.

For a link-based metric, those mentioned by Solmaz \textit{et al.} \cite{solmaz2017modeling} such as node density and intercontact time were difficult to apply in our dataset. We then considered Origin-Destination (OD) matrices, which have been commonly used in literature for analyzing flows between locations. For example, they have been used by Zhou \textit{et al.} \cite{zhou2020understanding} and Koh \textit{et al.} \cite{koh2020multiple} to illustrate the probability of human flows between different fixed nodes. However, we believe that they can be used to describe an individual's probability of motion between different locations as well, much like the links of a Markov chain model, which has been shown to have relatively high prediction accuracy \cite{lu2013approaching} for trajectories. Based on this, we propose a method of extracting OD matrix features in Section IV-A-2.

\section{Data Collection and Preprocessing}
The overall data flow of this paper is summarized in Fig.~\ref{fig:flowchart}. This section deals with the Data Collection and the Preprocessing parts. 

Timestamped GPS data was collected through a user-installed smartphone application that runs in the background and collects GPS data adaptively. When moving, data is recorded about once every minute, whereas when the device is still or not moving, the data is recorded about once every five minutes. Data collection was carried out over a range of different periods of time for different users. Usable data was selected with the criteria of at least one month of valid data accounting for at least 50\% of the recording duration for each user, resulting in the data from a total of 73 users being selected for use. Although the sample size is relatively small, it is due to strict criteria to ensure quality of the data used. Additionally, for this paper, we focus mainly on the framework, which can be extended to other datasets with larger sample size in the future. 

Each detected point of the data consists of a latitude, longitude, start time, and end time. For each user, individual points at similar coordinates were clustered together using a validation based stay point detection algorithm \cite{lau2017extracting} to identify points of interest (POIs). Home and Work locations for each user are then detected from this set of POIs using frequency and stay duration given the time of day, roughly based on a Monday to Friday workweek within standard office and retail hours. Taking into consideration that there may be differences in mobility patterns on Workdays and off days for the Working population, the POI data was then separated into Workday data and Offday data. Workdays are taken to be days when the user was detected at their Work location. Some of the users who are not detected in a fixed 'Work' location during those standard hours are considered as Nonworking users and all of their data is considered to lie in the Offday category.

Next, each POI is manually labeled by its proximity to the nearest location with a specific type out of ten different categories. If it is more than 400m away from any location with known POI types, it is left unlabeled. The labels are as shown in Table~\ref{tab:table1}.

\begin{table}[t]
	\def\arraystretch{1.5}
	\centering
	\caption{Labels for the different POI types considered in the dataset. }
	\begin{tabularx}{\columnwidth}{|s|b|}
		\hline
		\textbf{Label} & \textbf{Description}\\
		\hline
		Attraction & Places that tourists tend to visit \\
		\hline
		Healthcare & Hospitals, clinics etc. \\
		\hline
		Neighborhood Center & Community clubs, hawker centers, markets, etc.\\
		\hline
		Park & Public parks and gardens \\
		\hline
		Places of Worship & Temples, mosques, churches, etc. \\
		\hline
		Playground & Playgrounds\\
		\hline
		Recreational & Places that locals tend to visit for leisure\\
		\hline
		Shopping Mall & Shopping malls\\
		\hline
		Transportation & Train stations, bus interchanges, etc.\\
		\hline
		Residential & Condominiums, public housing estates, etc.\\
		\hline
	\end{tabularx}
	\label{tab:table1}
\end{table}

Finally, to minimize the impact of the GPS inaccuracy, the data points were then assigned to specific areas called subzones, which are small sections of planning areas delineated by the Urban Redevelopment Authority of Singapore for statistical purposes\cite{singstat2019}. These subzones were used in the extraction of clustering features, which can be found in Section IV-A-2.

\section{Proposed Clustering Methodology}
This section will go into details of the feature extraction and clustering processes. These processes differ slightly between the Workday and the Offday datasets. The aim of clustering these users is to find common types of users based on their mobility patterns, and thus possibly derive insights into common mobility patterns.

\subsection{Feature Extraction}
For the purposes of clustering, we extract two main types of features from each user. One is derived from a proposed metric, Daily Characteristic Distance (DCD), while the other is derived from the Origin-Destination (OD) matrix of each user's individual trips.

\subsubsection{Daily Characteristic Distance}
We are interested in mobility patterns for different users in the dataset. As different users have different Home and Work coordinates, it is imperative to find a mobility metric that enables us to compare different users despite this spatial restriction. One such metric in the literature to quantify the mobility of individuals is the radius of gyration, which considers distances from the center of mass of a trajectory and is thus user-dependent. The radius of gyration $r_g$ of a user $a$ from the start of their dataset up to a certain time $t$ was proposed by Gonzalez \textit{et al.} \cite{gonzalez2008understanding} and expressed by Equation~\ref{eq:originalrg}:
\begin{equation}
	r_g^a(t) = \sqrt{\frac{1}{n_c^a(t)}\sum_{i=1}^{n_c^a}(\vec{r}_i^a - \vec{r}_{cm}^a)^2}
	\label{eq:originalrg}
\end{equation} 
\noindent where $\vec{r_i^a}$ represents the $i = 1,...,n_c^a(t)$ positions recorded for user $a$ and $\vec{r_{cm}^a} = \frac{1}{n_c^a(t)}\sum_{i=1}^{n_c^a}\vec{r_i^a}$ is the center of mass of the trajectory. 

\begin{figure}
	\centering
	\includegraphics[width=\linewidth]{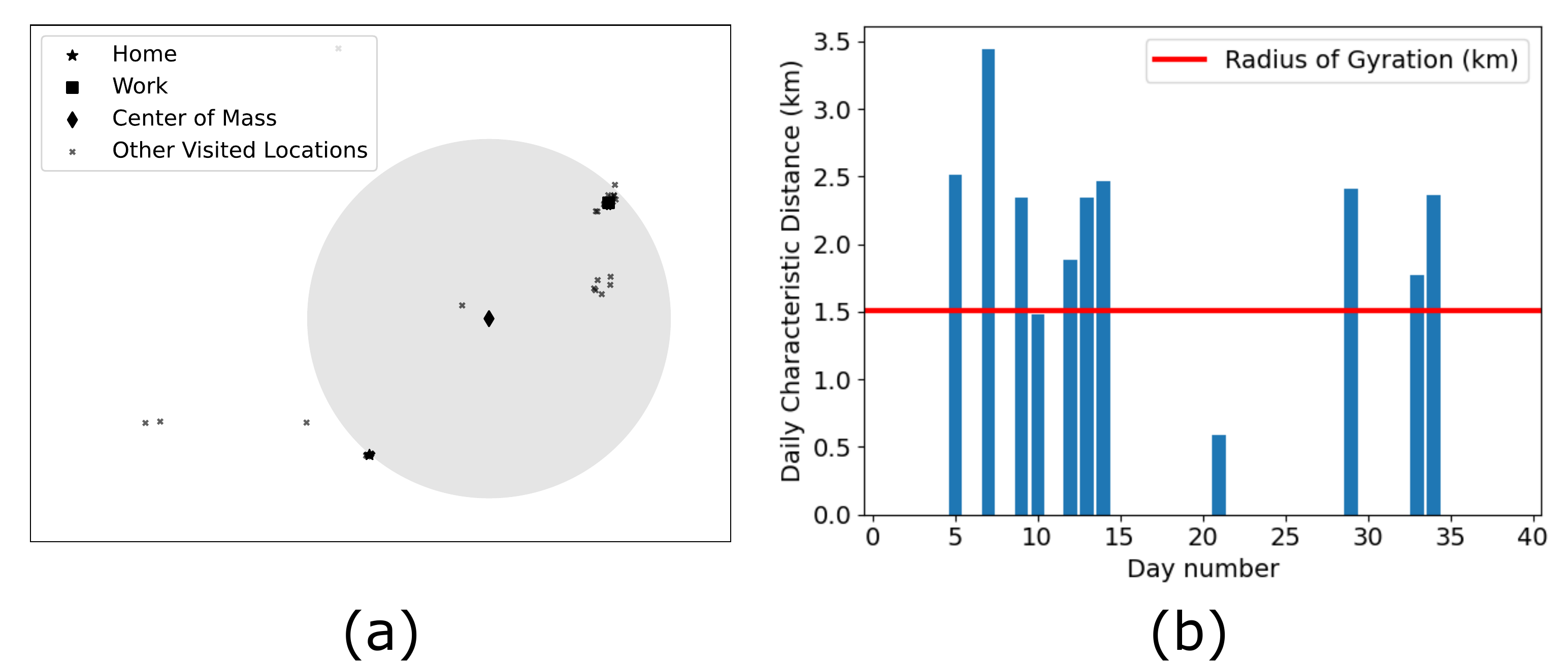}
	\caption{\textit{(a) Illustration of radius of gyration. The shaded gray circle is centered at the computed center of mass, with a radius equal to the computed radius of gyration. (b) Comparison between radius of gyration (single value, red line) and proposed Daily Characteristic Distance (set of values, bar plot) over the same time period.}}
	\label{fig:rgdcd}
\end{figure}

\begin{figure*}[htbp]
	\centering
	\includegraphics[width=\linewidth]{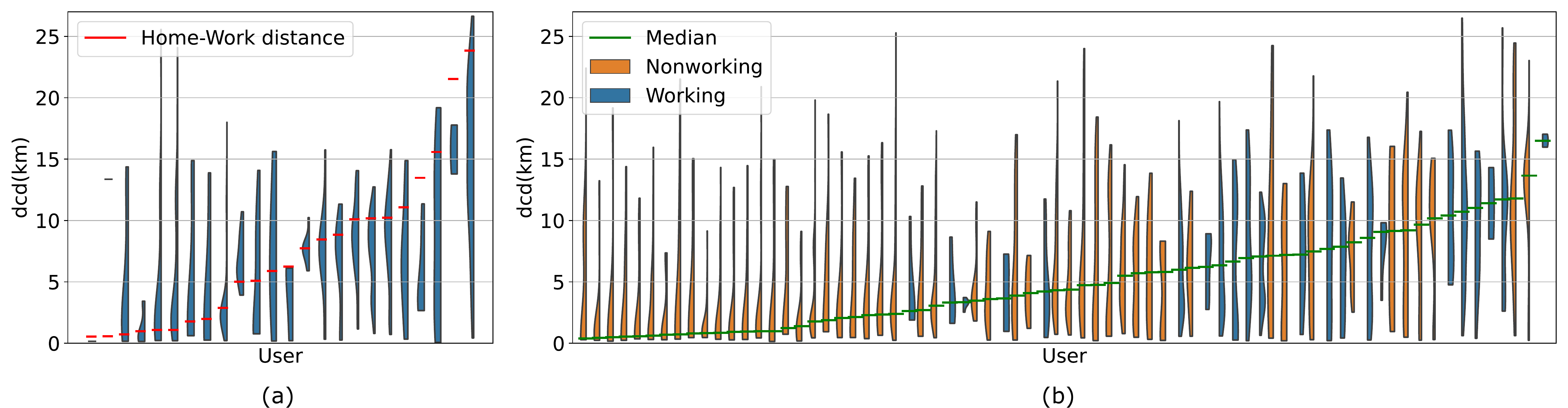}
	\caption{\textit{Violinplots illustrating the DCD distributions of users on (a) Workdays, consisting of only Working users, and (b) Offdays, consisting of both Working users on Offdays and Nonworking users on all days. (a) shows a moderately high correlation between DCD peaks and Home-Work distance of each user, while (b) shows a higher density of Working users with higher median DCD.}}
	\label{fig:dcdoverall}
\end{figure*}

For the purpose of comparison between different users in our dataset, the time $t$ in the above equation is taken to be the entire duration of each user's dataset, as the duration varies between users. Thus, the value of $n_c^a(t)$, which originally refers to the number of recorded positions of user $a$ up to time $t$,  becomes the total number of locations $N^a$ visited by user $a$ and the time dependency is removed. The simplified equation is as shown in Equation~\ref{eq:simplified}.
\begin{equation}
	r_g^a = \sqrt{\frac{1}{N^a}\sum_{i=1}^{N^a}(\vec{r}_i^a - \vec{r}_{cm}^a)^2}
	\label{eq:simplified}
\end{equation} 
An illustration for this metric is shown in Fig.~\ref{fig:rgdcd}(a) for a user with a 40-day dataset. As expected, the center of mass lies between the Home and Work location, as those locations are visited with a higher frequency than other locations.

However, there are some things that can be added to this metric, which it currently lacks. Firstly, as our dataset has labels for Home locations of each user, we can add more meaning to the metric by using the Home location of each user as the reference point for distance calculations instead of the computed center of mass of the user's visited locations. This will allow us to know the characteristic distance that the user travels from their Home, which may have more physical meaning than a computed center of mass of their trajectory. Secondly, the radius of gyration metric currently produces one value per user. We propose to break down the dataset into days and compute a value for each individual day, thus obtaining a distribution of the daily distances traveled by the user. Each day's characteristic distance may be affected by whether or not the user went to work on that day, which is part of what we want to investigate. Lastly, we want to investigate the locations that the users visit outside of Home and Work. Therefore, we manually negate the contribution of the Work and Home locations in the calculations by setting their distances to zero and removing the count of Home and Work visits from the total value of $N^a$. As our proposed new metric computes a characteristic distance for each day of the dataset, we call it Daily Characteristic Distance (DCD).

The DCD for day $d$ of the dataset is given by Equation~\ref{eq:dcd}: 
\begin{equation}
	DCD_d = \sqrt{\frac{1}{n_d}\sum_{i=1}^{n_d}f_{id} \times (\vec{r}_{id} - \vec{r}_{home})^2}
	\label{eq:dcd}
\end{equation} 
\noindent where $n_d$ is the number of unique POIs that the user traveled to on that day, $f_{id}$ is the number of times the user traveled to location i where $i = 1,...,n_d$ on that day, and $\vec{r}_{home}$ is given by the mean coordinates of the Home location of the user. Fig.~\ref{fig:rgdcd}(b) illustrates the difference between radius of gyration (one value per user) and DCD (a set of values per user). The days without bars have a value of zero, indicating that on those days the user traveled directly between Home and Work without visiting any other location.

The obtained DCD distributions of all users are plotted in Fig.~\ref{fig:dcdoverall}. The Workday data of Working users is plotted in Fig.~\ref{fig:dcdoverall}(a), while the Offday data (consisting of Working users during Offdays and Nonworking users on all days) is plotted in Fig.~\ref{fig:dcdoverall}(b). In Fig.~\ref{fig:dcdoverall}(a), we have sorted the distributions in ascending order of Home-Work distance of each user. From this, we can see that there generally seems to be a relationship between the Home-Work distance of the users and the location of the peaks of their DCD distributions. We calculated the Pearson's R-value between each user's Home-Work distance and the median of their DCD distribution and found that there was a moderately high R-value of 0.746, with a p-value of $2.90\mathrm{e}{-5}$.

For the Offday data, since there is no second location outside of Home that was fixed for every user, we could not apply this method. Instead, we have sorted the distributions in ascending order of their median point and colored the distributions according to whether the user is a Working or Nonworking user. From Fig.~\ref{fig:dcdoverall}(b), we can see that there is a higher concentration of Working users (blue) at the side with higher median DCD. This may indicate that Working users tend to visit locations at further distances from their homes as compared to Nonworking users.

\begin{figure}[htbp]
	\centering
	\includegraphics[width=\linewidth]{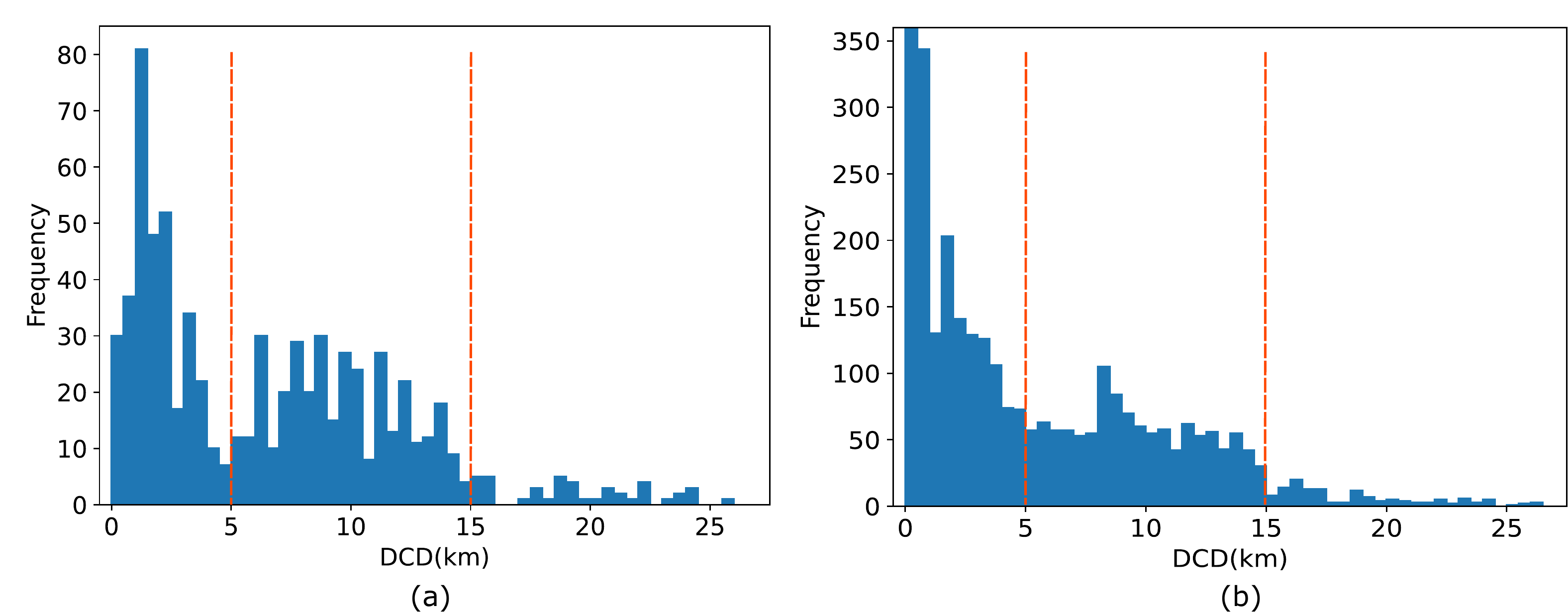}
	\caption{\textit{Histograms of the number of days within the whole dataset of (a) DCD value on Workdays, and (b) DCD value on Offdays. Note that (b) has been cropped vertically to show greater detail - the leftmost bar has an actual value of 3312, of which 3137 have a value of 0.}}
	\label{fig:dcdhist}
\end{figure}

To use this new metric DCD as a clustering feature, we first break down all the data for all users into Workday and Offday data. We then separately compute the DCD values for each day and plot separate histograms of the DCD values over all relevant users on Workdays and Offdays, as shown in Fig.~\ref{fig:dcdhist}. From these histograms, we visually obtain the four thresholds of 0km (Home and/or Work only), 0 to 5km, 5 to 15km, and $>$15km by observing suitable valleys.

Each user's data is then broken down into Workday (if applicable) and Offday data. For each type of data, we calculate the percentage of DCD values that fall within each of the determined thresholds. This gives us four features for each type of data that add up to 1.0. An example of the features for one user, User 2, is shown in Table~\ref{tab:dcdfeat}.

\begin{table}[ht]
	\def\arraystretch{1.5}
	\centering
	\caption{Example of the four DCD features for Workday and Offday data.}
	\begin{tabular}{|c|c|c|c|c|}
		\hline
		\textbf{Day Type} & \textbf{Home/Work} & \textbf{0-5km} & \textbf{5-15km} & \textbf{>15km}\\
		\hline
		Workday & 0.60 & 0.23 & 0.15 & 0.02\\
		\hline
		Offday & 0.38 & 0.16 & 0.41 & 0.05\\
		\hline
	\end{tabular}
	\label{tab:dcdfeat}
\end{table}

These four features will be used in conjunction with the 16 features derived from the Origin-Destination Matrix, explained below:\\

\subsubsection{Origin-Destination Matrix}
\begin{figure}
	\centering
	\includegraphics[width=\linewidth]{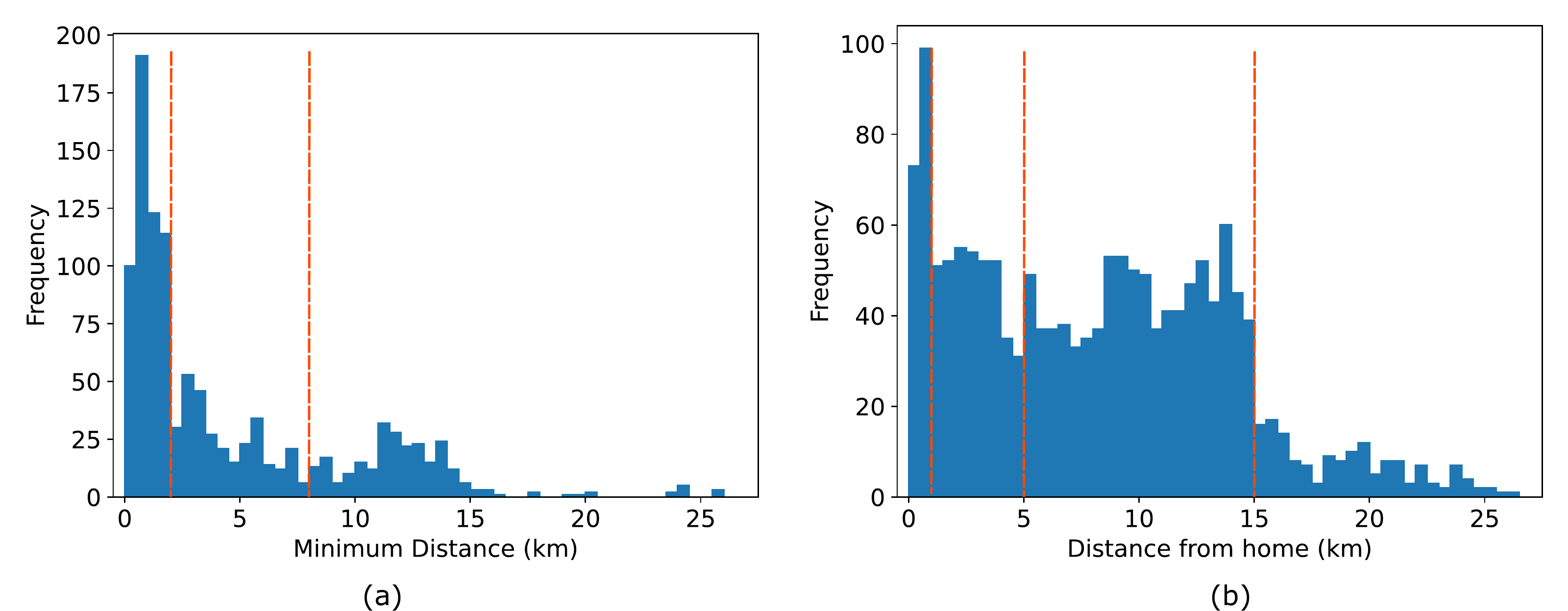}
	\caption{\textit{Histograms of the number of POIs visited over the whole dataset of (a) minimum distance between Home and Work to that location on Workdays, and (b) distance from home to that location on Offdays.}}
	\label{fig:disthist}
\end{figure}

From each user's trajectory, we can extract distances from each user's Home (and Work location if applicable) to the other POIs that the user visits. For Offdays, we can simply use the distance from Home to that POI as there is no other location that is common to all users. However, there is an additional important location for Workdays, which is the Work location of the user. Therefore, on Workdays, the distance value of each POI, referred to as minimum distance, is taken as the minimum of the distances between the POI and the user's Home location and between the POI and the user's Work location. This is to detect any specific locations that users may go to, that is not nearby to either their Home location or Work location and hence ``out of the way" from the user's point of view. After extracting these distances separately from the Workday and Offday data, the corresponding histograms are plotted. These can be seen in Fig.~\ref{fig:disthist}. We obtain the thresholds visually by selecting suitable valleys in the histograms. The thresholds for Workdays are 0km (direct trips between Home and Work), 0-2km, 2-8km, and $>$8km, while the thresholds for Offdays are 0-1km, 1-5km, 5-15km, and $>$15km.

After getting these thresholds, the trips made by a user are now categorized based on these thresholds. We are interested in the combinations of movements that users make from threshold to threshold, and whether these will be a significant distinguishing factor between different users' mobility patterns. Taking an example of a user with Workday data, a trip consists of going from location A to location B, where threshold A is on the row of the matrix and threshold B is on the column of the matrix. If A is located within 0-2km and B is located within 2-8km, the number corresponding to the ``0-2km" row and the ``2-8km" column will be increased by 1. After the trips are all counted for a user, the matrix is normalized by the total number of trips counted for that user, such that all 16 elements of this matrix add up to 1.0. This is to make the data comparable between different users. An example of the resulting matrix using the Workday thresholds can be seen in Table~\ref{tab:odfeat}. The Offday matrix and features are computed similarly.

We do not count trips occurring on different calendar dates (i.e. from the last POI on one day to the first POI the next morning), and we also do not count trips that occur within the same subzone (e.g. Home to Home). 

\begin{table}[ht]
	\def\arraystretch{1.5}
	\centering
	\caption{Example of the 16 OD matrix features for Workday and Offday data.}
	\vspace{2ex}
	(a) Workday features\\
	\vspace{2ex}
	\begin{tabular}{|c|c|c|c|c|}
		\hline
		\textbf{Threshold} & \textbf{Home/Work} & \textbf{0-2km} & \textbf{2-8km} & \textbf{>8km}\\
		\hline
		\textbf{Home/Work} & 0.68 & 0.08 & 0.04 & 0.04\\
		\hline
		\textbf{0-2km} & 0.08 & 0.00 & 0.00 & 0.00\\
		\hline
		\textbf{2-8km} & 0.04 & 0.00 & 0.00 & 0.00\\
		\hline
		\textbf{>8km} & 0.03 & 0.00 & 0.00 & 0.01\\
		\hline
	\end{tabular}\\
	\vspace{3ex}
	(b) Offday features\\
	\vspace{2ex}
	\begin{tabular}{|c|c|c|c|c|}
		\hline
		\textbf{Threshold} & \textbf{0-1km} & \textbf{1-5km} & \textbf{5-15km} & \textbf{>15km}\\
		\hline
		\textbf{0-1km} & 0.00 & 0.08 & 0.23 & 0.03\\
		\hline
		\textbf{1-5km} & 0.08 & 0.03 & 0.00 & 0.00\\
		\hline
		\textbf{5-15km} & 0.19 & 0.04 & 0.29 & 0.00\\
		\hline
		\textbf{>15km} & 0.02 & 0.00 & 0.01 & 0.00\\
		\hline
	\end{tabular}
	\label{tab:odfeat}
\end{table}

These 16 features are then concatenated with the four features from the above DCD calculations to form the 20 features used in clustering, which will be discussed in detail in the next subsection.

\subsection{Proposed Clustering Process}
After extracting the features for each user as in the above subsection, we performed some initial test clustering, using Euclidean distance as a distance measure between users, to obtain the sum-of-squared errors (SSE) plot. This plot was used to decide on the number of clusters, $k$, to be used as the input parameter for $k$-means clustering\cite{lloyd1982least}. 

\begin{figure}
	\centering
	\includegraphics[width=\linewidth]{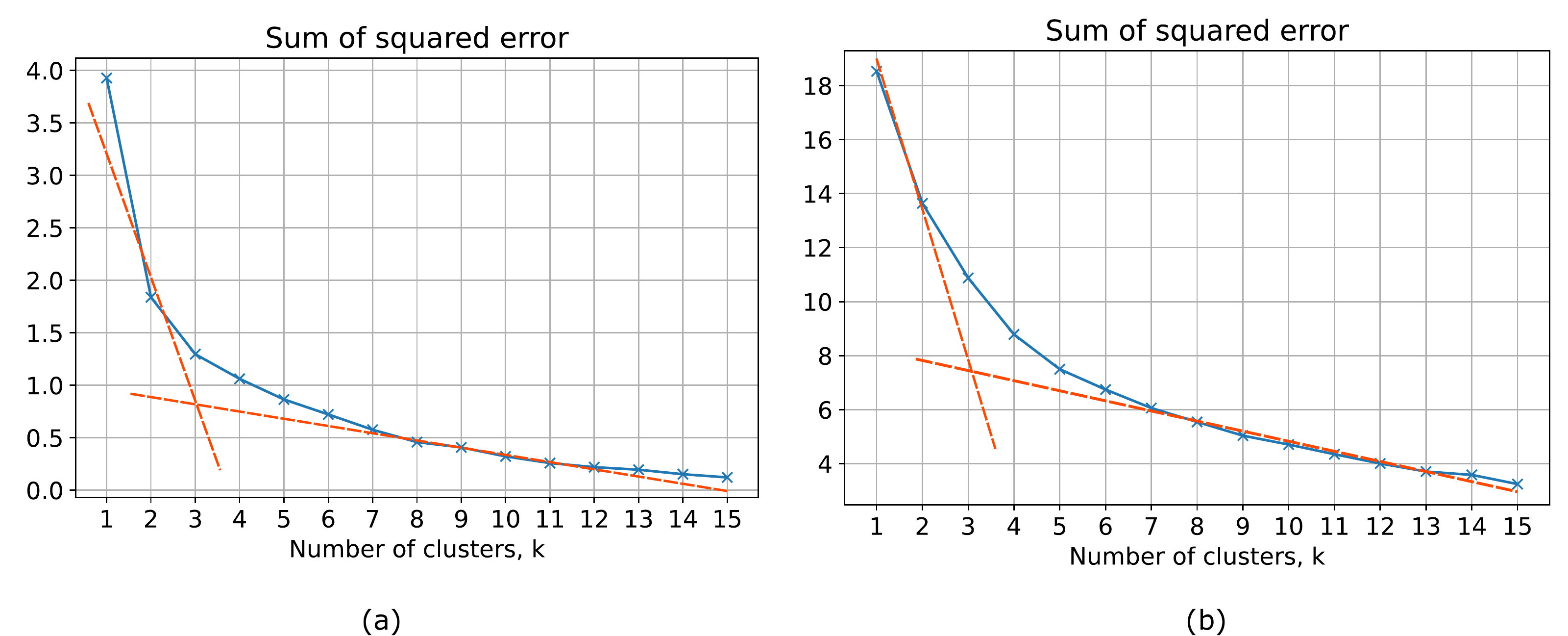}
	\caption{\textit{SSE plots used to derive (a) the optimal number of clusters for Workdays and (b) the optimal number of clusters for Offdays. Both plots indicate 3 as a suitable value for $k$, the number of clusters.}}
	\label{fig:sse1}
\end{figure}

The SSE plot measures the sum of all squared errors from the clustered points to their respective cluster centers after being grouped using each value of $k$. As the value of $k$ increases, the SSE naturally decreases, but a good value for $k$ would be one located at the `elbow' of the plot, just before the decrease in SSE becomes less than proportionate to the increase in $k$. The SSE plots for our dataset can be seen in Fig.~\ref{fig:sse1}, where Fig.~\ref{fig:sse1}(a) shows the plot using the data from Workdays, while Fig.~\ref{fig:sse1}(b) shows the plot using the data from Offdays. From both SSE plots, the `elbow' of the plot indicates that a good value of $k$ to use would be $k$ = 3. The detailed results are plotted in the following sections, with the Workday results presented first before Offday results.

\begin{figure*}[ht]
	\centering
	\includegraphics[width=\linewidth]{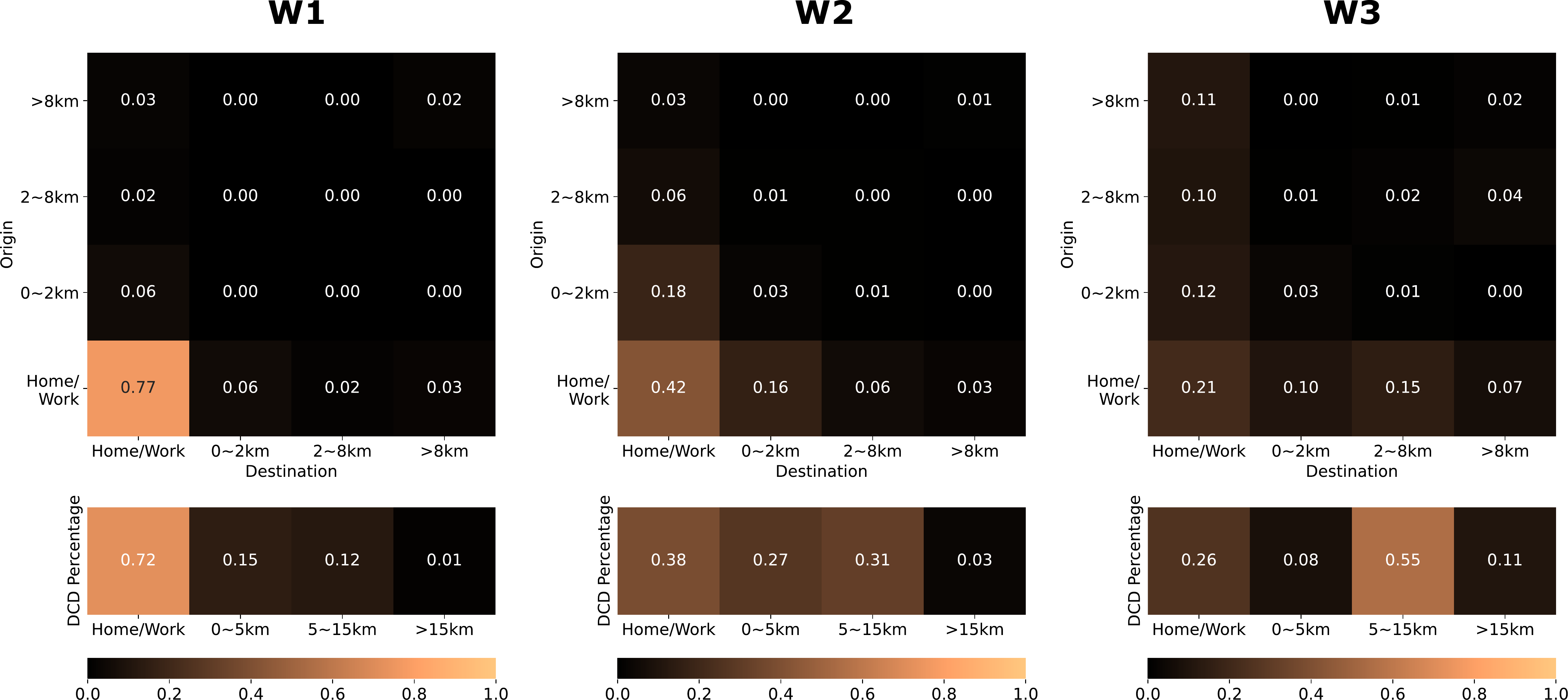}
	\caption{\textit{Centroid values of the three clusters obtained from clustering Workday data. Cluster W1 has the highest percentage of trips directly between Home and Work, as well as the highest percentage of days spent only at Home or Work. The other two clusters W2 and W3 are in descending order of percentage of trips directly between Home and Work.}}
	\label{fig:centroidwork}
\end{figure*}

\section{Workday Clustering and Analysis}
This section focuses on the results obtained from clustering the Workday data of working users. The eventual aim of this clustering is to separate users into different clusters based on their workday data. Further analysis is then performed on the identified clusters, which consists of the DCD violinplots for each user in each cluster, as well as user commonality and average frequency heatmaps, which are explained in detail later on. The same process will be repeated for the Offday data in the next section.

\subsection{Clustering Results - Centroid Values}

The centroid values of each cluster are shown in Fig.~\ref{fig:centroidwork}. These values represent the average percentage of trips within each threshold (for the first 16 values in the O-D matrix) and the last 4 values represent the percentage of days for each user that have DCD values within each of the 4 distance thresholds, as described in Section IV-A. The clusters are named W1, W2, and W3 respectively (W stands for Workday). Visually, it seems that the clusters are separated mainly based on the percentage of Home/Work trips out of the total number of trips taken by the user, with cluster W1 having the highest average percentage of direct Home-Work trips at 77\% followed by W2 with 42\% and W3 with 21\%. The DCD features below each OD matrix show that there are similar average percentages of Home/Work Only trips and Home/Work Only days.

Users from Cluster W1 have a large majority - on average 72\% - of their Workdays where they do not visit any other locations. The average percentage of their days spent with DCD at each distance threshold decreases with increasing distance.

Looking at Cluster W2, the DCD features are roughly evenly spread across the first three distance thresholds. Since there is a higher value of DCD being within 5-15km as compared to 0-5km, while the percentage of trips from the OD matrix indicate a higher emphasis on minimum distance between 0-2km, it is likely that some of the locations, which are 5-15km from their Home, are actually within 0-2km of their workplace, leading to a lower value for minimum distance.

For Cluster W3, the DCD features have a surprisingly high average value of 55\% in the 5-15 km threshold as compared to 12\% and 31\% the other two clusters. It also has a much lower average value of 8\% in the 0-5km threshold, as compared to 15\% and 27\% in the other clusters. As the average percentage of Home/Work direct trips from the OD matrix is also quite low at 21\%, it can be interpreted that the users in this cluster usually travel quite far from their Home and Work locations.

Overall, the clusters can be described as mainly Home/Work Only (W1), frequent short trips in terms of Minimum Distance (W2), and mostly longer trips (W3).

\begin{figure*}[htbp]
	\centering
	\includegraphics[width=\linewidth]{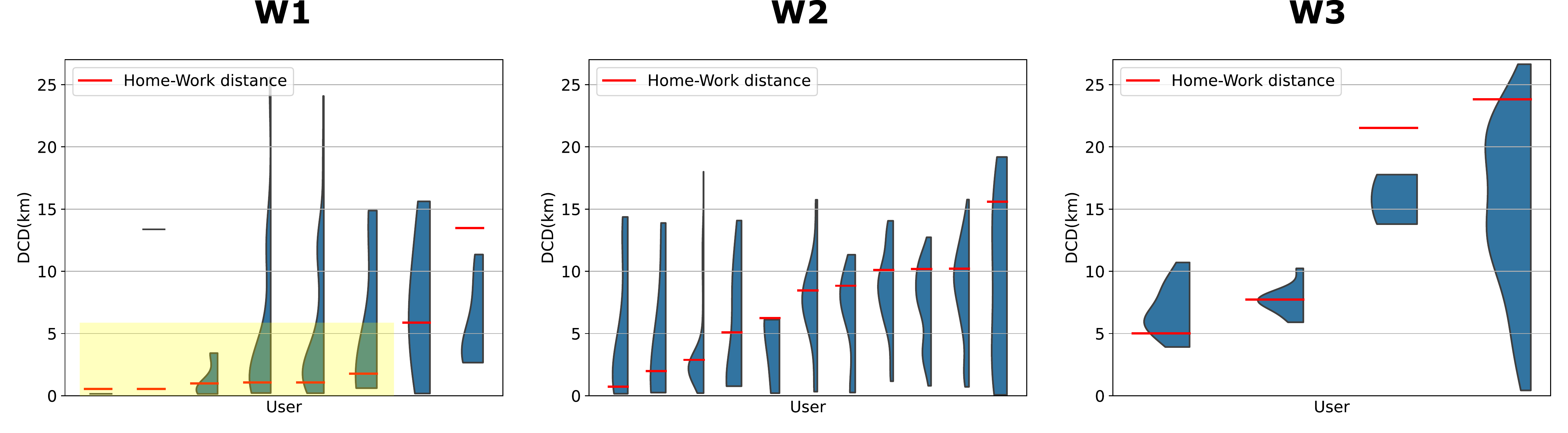}
	\caption{\textit{Violinplots illustrating each user's DCD distribution within each cluster on Workdays.}}
	\label{fig:violinswork}
\end{figure*}

\begin{figure*}[htbp]
	\centering
	\includegraphics[width=\linewidth]{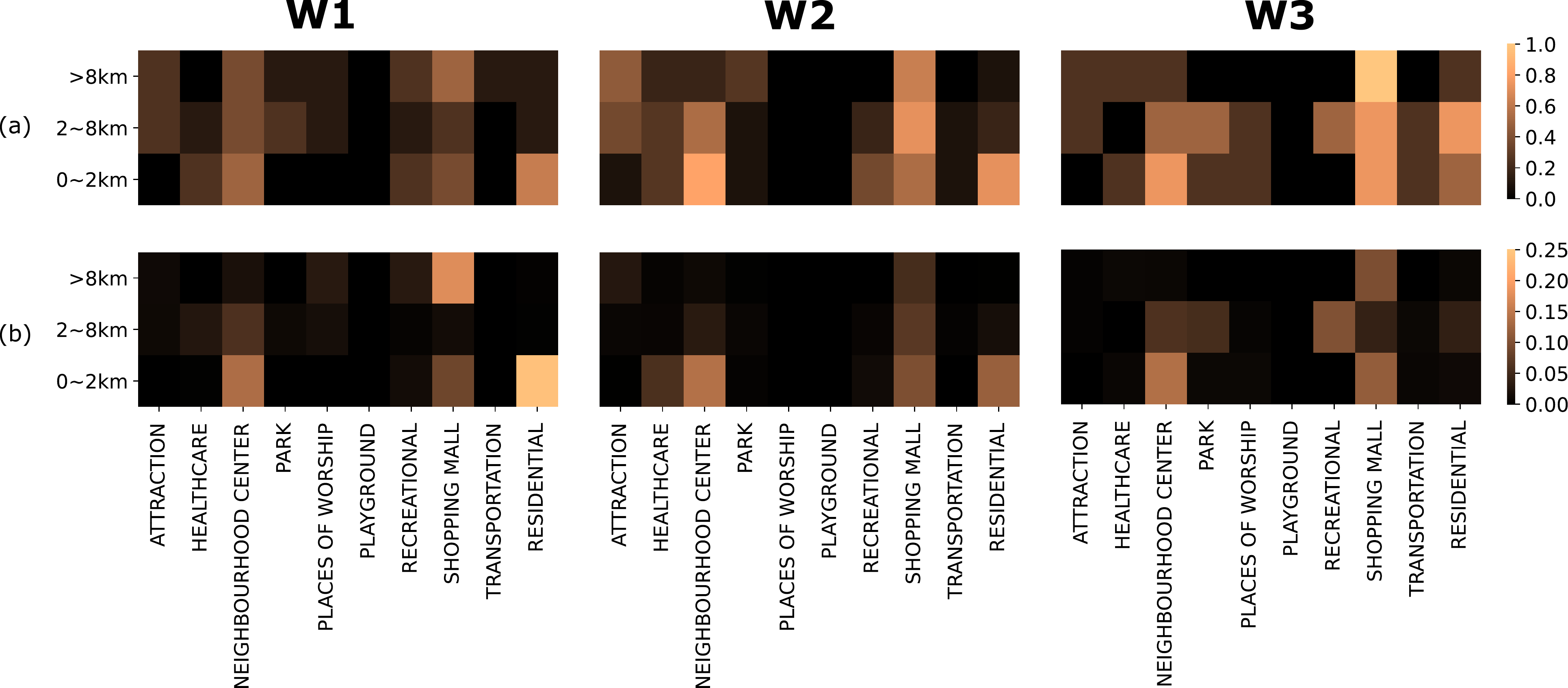}
	\caption{\textit{Heatmaps for each of the three Workday clusters showing (a) User Commonality and (b) Average Frequency. The colormap scales for (b) are narrowed to 0.25 to better show the contrast between the different squares.}}
	\label{fig:furtherwork}
\end{figure*}

\subsection{Cluster Analysis - DCD Violinplots}

Fig.~\ref{fig:violinswork} shows the DCD violinplot for each individual user in each cluster, sorted in ascending order of their Home-Work distance. These violinplots do not show the percentage of days spent only between Work and Home, as we are interested in the POIs that are not Home and not Work. From this figure, we observe from the yellow highlighted portion that most of the users in cluster W1 have a low Home-Work distance, below 5km. This may be a factor in these users having the highest percentage of direct trips between Home and Work out of the three clusters on Workdays. The users in cluster W2 have Home-Work distances in the middle range, and usually the peaks of their DCD distributions are located close to the Home-Work distances. This is also reflected in their OD matrix, in which this cluster has the highest percentage of trips within 0-2km of either their Home or Work location out of the three clusters. For Cluster W3, two out of the four users have a large Home-Work distance of over 20km. Three out of the four users have DCD peaks near their Home-Work distance, but those are not reflected in the centroid OD matrix, perhaps because they travel to other places that are the same distance from their Home as well as their Work location. These DCD plots are in agreement with the DCD features for Cluster W3, as the bulk of their DCD distributions are located within the 5-15km range.

\subsection{Cluster Analysis - User Commonality and Average Frequency}

The next two parts of cluster analysis, what we will call User Commonality and Average Frequency, are shown in  Fig.~\ref{fig:furtherwork} (a) and (b) respectively. Both of these types of analysis use the same distance thresholds for minimum distance that were used for the OD matrix features, broken down into each of the ten different POI categories that were labeled in the data. To represent User Commonality, each square in Fig.~\ref{fig:furtherwork}(a) shows the percentage of users within the cluster who fulfilled each minimum distance and POI label combination at least once in their trajectory. The aim of this is to see whether there is any minimum distance and POI label combination that is favored by the users in each cluster. The value of each heatmap square $u_{jk}$, in row $j$ and column $k$, is given by Equation~\ref{eq:usercomm}: 
\begin{equation}
	u_{jk} = \frac{n_{jk}}{n_c}
	\label{eq:usercomm}
\end{equation}
\noindent where $n_{jk}$ is the number of users within the cluster who visited a POI at distance threshold $j$ with label $k$, and $n_c$ is the total number of users in that cluster.

Meanwhile, Fig.~\ref{fig:furtherwork}(b) shows the Average Frequency, taken as a percentage of the user's total trips and averaged over all users in the cluster, of each minimum distance and POI label combination. The value of each heatmap square $f_{jk}$, in row $j$ and column $k$, is given by Equations~\ref{eq:singlefreq} and \ref{eq:clusterfreq}:
\begin{equation}
	P_{ijk} = \frac{p_{ijk}}{p_i}
	\label{eq:singlefreq}
\end{equation}
\begin{equation}
	f_{jk} = \frac{\sum_{i=1}^{n_c}P_{ijk}}{n_c}
	\label{eq:clusterfreq}
\end{equation}
\noindent where $P_{ijk}$ is the percentage frequency of user $i$ visiting a POI at distance threshold $j$ and with label $k$, $p_{ijk}$ is the number of POIs that user $i$ visited at distance threshold $j$ with label $k$, $p_i$ is the total number of labeled POIs visited by user $i$, and $n_c$ is the total number of users within the cluster.

The advantage of using these two analysis metrics is that they both make use of the data available within the clusters, and thus do not require an external source of ground truth, to highlight meaningful differences between the clusters that may not be apparent at first glance. 

From Fig.~\ref{fig:furtherwork}(a), it can be seen that there is no single distance threshold and POI label combination that is visited by 100\% of the users in each cluster, except for Shopping Malls at a minimum distance of $>$8km for Cluster W3. However, quite a high percentage of users in the other two clusters visit this distance threshold/POI label combination as well. Other common combinations that appear in all three clusters are: Neighborhood Center, Shopping Mall, and Residential, all within the 0-2km threshold. The distinguishing features of the clusters can be summarized as follows: Cluster W1 has a visible percentage at Recreational at $>$8km minimum distance as compared to the others. More of the users in cluster W2 visit Attractions at a minimum distance of larger than 8km as compared to the other clusters. Users in Cluster W3 have a higher frequency of trips to various places over a variety of distance thresholds, such as Parks, Recreational, and Residential areas within the 2-8km range, as well as Healthcare, Neighbourhood Center, and Shopping Mall in the >8km range.

From Fig.~\ref{fig:furtherwork}(b), it can be seen that the POI label with common frequency among the three clusters is Neighborhood Center at 0-2km. Cluster W1 has highest Average Frequency at Residential POIs within 0-2km and Shopping Malls at $>$8km. In comparison to Cluster W2, which has the highest relative frequency of shopping mall trips at 0-2km, this indicates that the users in Cluster 1 may be more willing to go a further distance on their shopping trips. Cluster W2 has a visible frequency at Healthcare at 0-2km, something which is not seen in the other two clusters. The users in Cluster W2 may visit Healthcare locations more frequently, and it makes sense that they would primarily visit Healthcare locations that are nearer to either their Home or Work locations. Cluster W3 has a visible frequency at the Park and Recreational POIs within the 2-8km threshold, which is not observed in the Cluster W1 and Cluster W2. This may indicate that users in Cluster W3 make trips to areas related to leisure more frequently than the users in the other clusters.

Comparing the two parts of Fig.~\ref{fig:furtherwork}, we can see that although there are some distance and label combinations that have more users in each cluster that visit them, it does not necessarily mean that they visit them frequently. The label/distance combinations that are visited frequently are a subset of those that are visited commonly by users.

\section{Offday Clustering and Analysis}

This section describes the results obtained from clustering the Offday data of all users. The process is the same as the one used for the Workday data in Section V. The three clusters here are labeled O1, O2, and O3, with `O' standing for `Offday'.

\subsection{Clustering Results - Centroid Values}

\begin{figure*}[htbp]
	\centering
	\includegraphics[width=\linewidth]{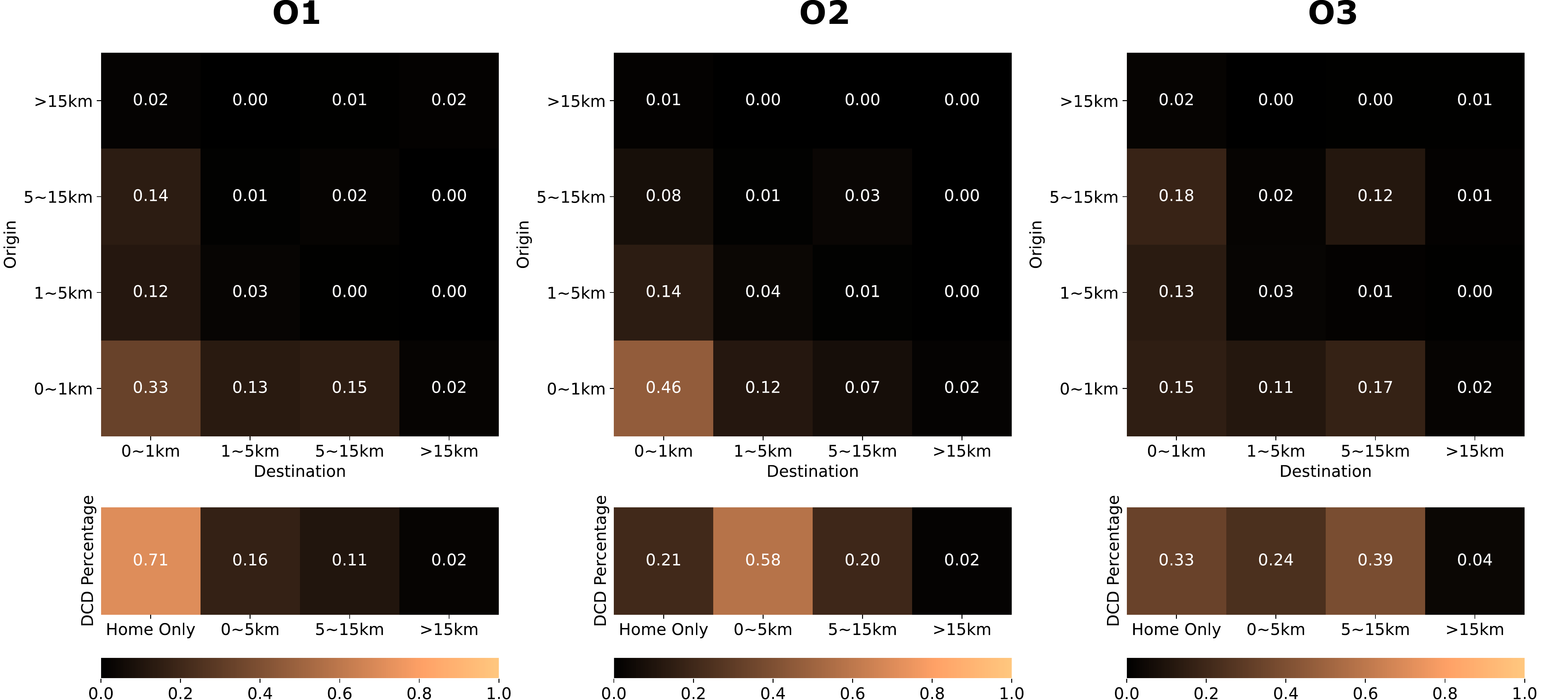}
	\caption{\textit{Centroid values of the three clusters obtained from clustering Offday data. Cluster (a) has the highest percentage of days spent at Home only, while cluster (b) has the highest percentage of days with DCD between 0 to 5 km, meaning they went to at least one other non-Home location. Cluster (c) has the highest percentage of days with DCD in the 5 to 15km range, indicating that they generally travel the furthest on Offdays.}}
	\label{fig:centroidoff}
\end{figure*}

\begin{figure*}[htbp]
	\centering
	\includegraphics[width=\linewidth]{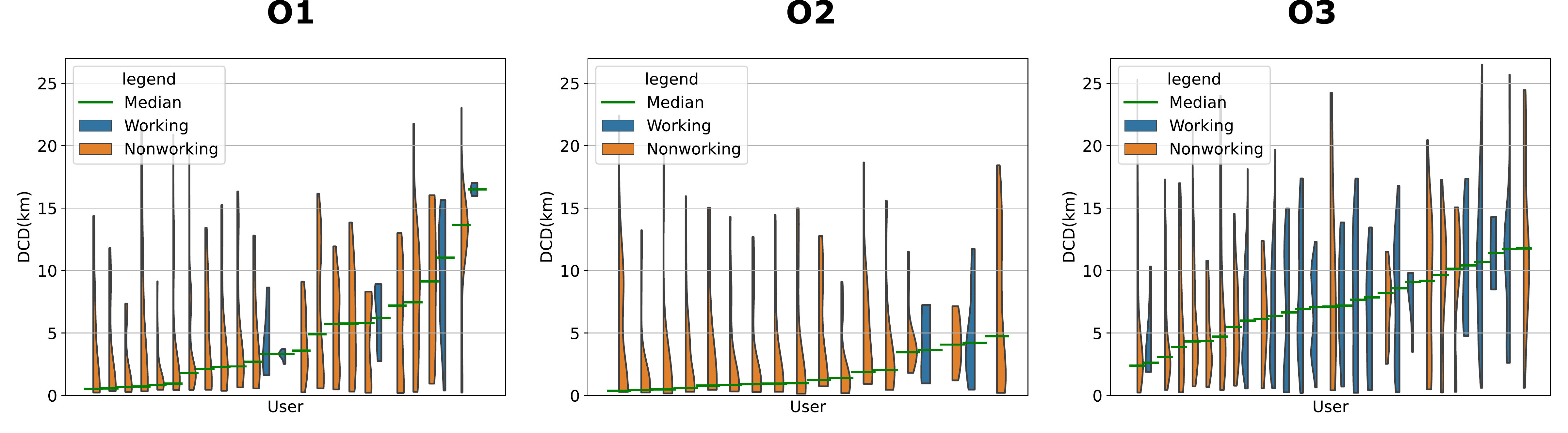}
	\caption{\textit{Violinplots illustrating each user's DCD distribution within each cluster on Offdays.}}
	\label{fig:violinsoff}
\end{figure*}

The values of each cluster's centroids are plotted in Fig.~\ref{fig:centroidoff}. We can observe that these clusters show similar trends to the Workday clusters in that there are those that stay mostly at Home Only (O1), those that make mostly short trips (O2), and those that make mostly longer trips (O3). The users in Cluster O1 spent 71\% of their Offdays only at their Home location. Cluster O2 users spent on average 21\% of their days at their Home location, and 58\% of their days have a DCD value of 0-5km. The average percentages for Cluster O3 are more evenly split between the Home Only and the first two distance categories, with the highest being 39\% of days with DCD values of 5-15km.

When observed together with each cluster's corresponding OD matrix, we see that Cluster O2 actually has the highest average percentage of trips within 0-1km at 46\% compared to 32\% for Cluster O1. Additionally, we see that the percentages of trips going between the 0-1km threshold and further thresholds is actually higher in Cluster O1 than Cluster O2. A possible reason for this could be that although the users in O1 stay at home only for more days than those in O2, they tend to travel further when they do go out, whereas those in O2 could go out on more days but stay within 0-1km for most of their trips. The users in cluster O3 seem to have more of a balance between staying at home and going out to near or further places.

\begin{figure*}[htbp]
	\centering
	\includegraphics[width=\linewidth]{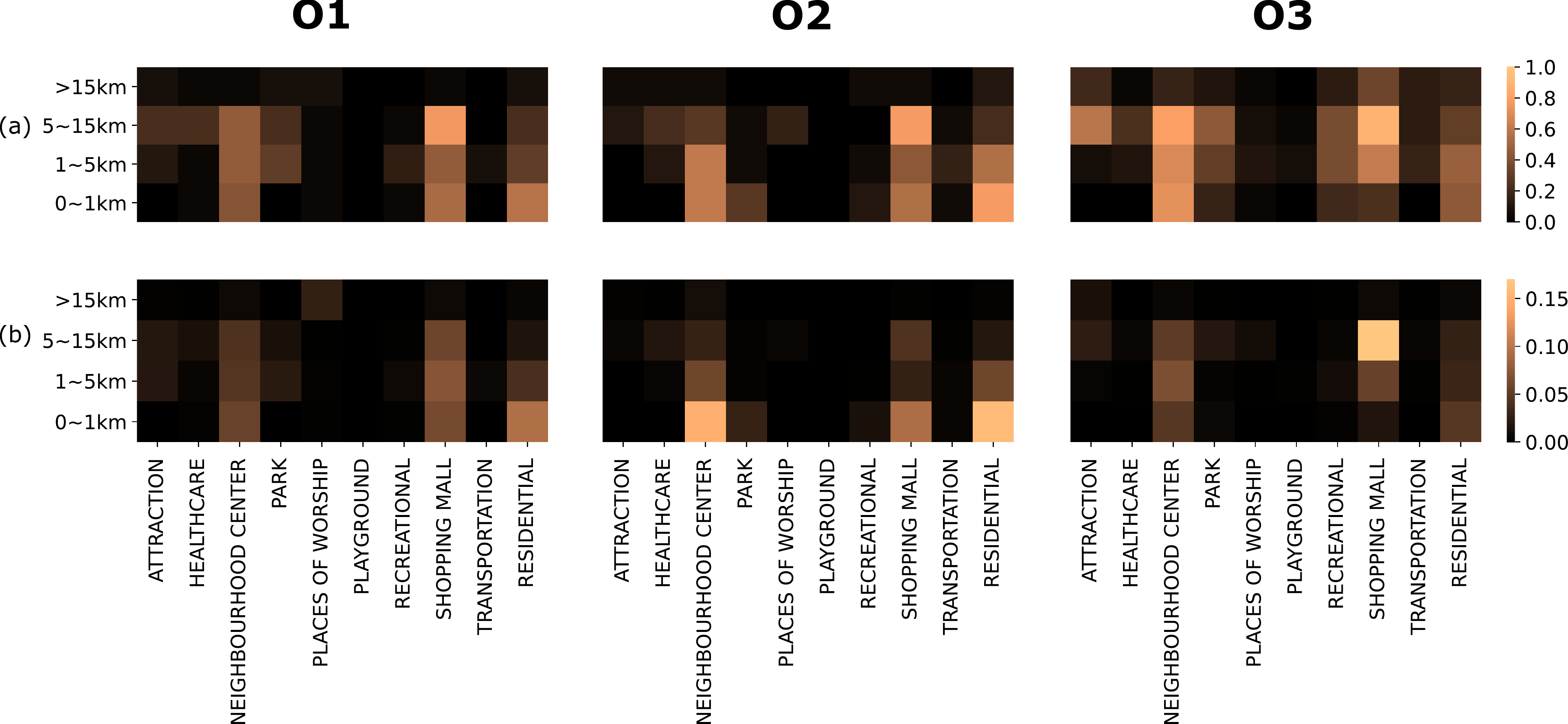}
	\caption{\textit{Heatmaps for each of the three Offday clusters showing (a) User Commonality and (b) Average Frequency. The colormap scales for (b) are narrowed to 0.17 to better show the contrast between the different squares.}}
	\label{fig:furtheroff}
\end{figure*}

\subsection{Cluster Analysis - DCD Violinplots}
The violinplots representing the DCD distribution of each user within each cluster have been plotted in Fig.~\ref{fig:violinsoff}. Similarly to before, the Home Only days are not reflected on this plot as we are more interested on days in which the users do go out.

Cluster O1 and Cluster O2 both contain dominantly Nonworking users, while the bulk of the Working users are in Cluster O3. Qualitatively speaking, Cluster O1 seems to lie in the middle of Clusters O2 and O3. The median DCDs of the users in Cluster O2 are limited to the 0-5km range, which agrees with the DCD features observed in Fig.~\ref{fig:centroidoff} and further emphasizes that this group of users makes mostly short trips. Although the median values of Cluster O3 are not always higher than those in O2, the bulk of the DCD distributions for Cluster O3 lies above 5km, which is the distance threshold for longer trips in this case.

\subsection{Cluster Analysis - User Commonality and Average Frequency}

User Commonality and Average Frequency of each cluster is obtained as described earlier in Section V-C, and plotted in Fig.~\ref{fig:furtheroff}. From Fig~\ref{fig:furtheroff}(a), we observe that there are the same three main POI labels that are commonly visited by users, namely Neighborhood Center, Shopping Mall, and Residential. There is a higher percentage of users in Cluster O3 who visit Parks and Recreational areas between 1-15km, as well as visiting Attractions that are in the 5-15km range from their homes. This may imply that users who have a higher median DCD tend to visit a variety of locations on Offdays, which are also further from their Home locations.

Looking at Fig.~\ref{fig:furtheroff}(b), we see that for all three clusters, the three highest frequency labels are the same as the highest commonality labels. However, for Shopping Malls, the highest frequency distance threshold differs for each cluster. For Cluster O1, the frequency is higher for Shopping Malls in the 1-5km range. For Cluster O2, the frequency is concentrated at the 0-1km range for Shopping Malls and similarly for Neighborhood Center and Residential areas. We can infer that the frequent short trips for this cluster are mainly for the purpose of visiting locations with those three labels. For Cluster O3, the frequency is prominently concentrated at Shopping Malls in the 5-15km range, and the frequency for Residential areas is much lower than for the other two clusters. This implies that shopping malls are a common destination further away from their home and work for these users in Cluster O3.

\section{Conclusion}
In this paper, we investigated the differences between the GPS trajectory patterns of Workday and Offday data, as well as Working and Nonworking users. To do so, we proposed a new mobility metric based on radius of gyration, named Daily Characteristic Distance (DCD), to zoom in on the locations outside of Home and Work if applicable that the users visited. We discover that Working users' median DCD on Workdays is highly correlated to the distance between their Home and Work locations, and that Working users generally have a higher median DCD on Offdays as compared to Nonworking users. 

We then used features derived from DCD in conjunction with those derived from the users' Origin-Destination matrices to cluster the users in our dataset. We find that we can group users' mobility into three types for both Workdays and Offdays. The three types are mainly those that mainly stick to Home (and Work if applicable), those that make frequent short trips, and those that make longer trips. We also propose two new types of metric for cluster analysis, namely User Commonality and Average Frequency, to better assess the labels and distances of different locations that are favored by the users in different clusters. We discover that three main POI labels are favored regardless of cluster - Neighborhood Centers, Shopping Malls, and Residential areas, but the main differences between clusters are the distance thresholds of these POI labels, as well as the presence or absence of some other labels such as Attraction, Parks, and Recreational areas. Urban planners could use this framework on their own target datasets as a case study to discover the types of places that would be beneficial to locate nearby their intended residential environment. It is important to note that, while our proposed framework is general, the results that we have obtained are dependent on our data that we have gathered in Singapore, and thus results may differ widely if our framework is used on data from other countries.

Currently, our work examines the users' data and clusters them separately for Workdays and Offdays. There could be more insights to be drawn from linking both the Workday and Offday mobility features of the individual Working users together and examining the resulting features for new correlations. This could be a part of future work.

\bibliographystyle{IEEEtran}
\bibliography{bibliography}

\begin{IEEEbiography}[{\includegraphics[width=1in,height=1.25in,clip,keepaspectratio]{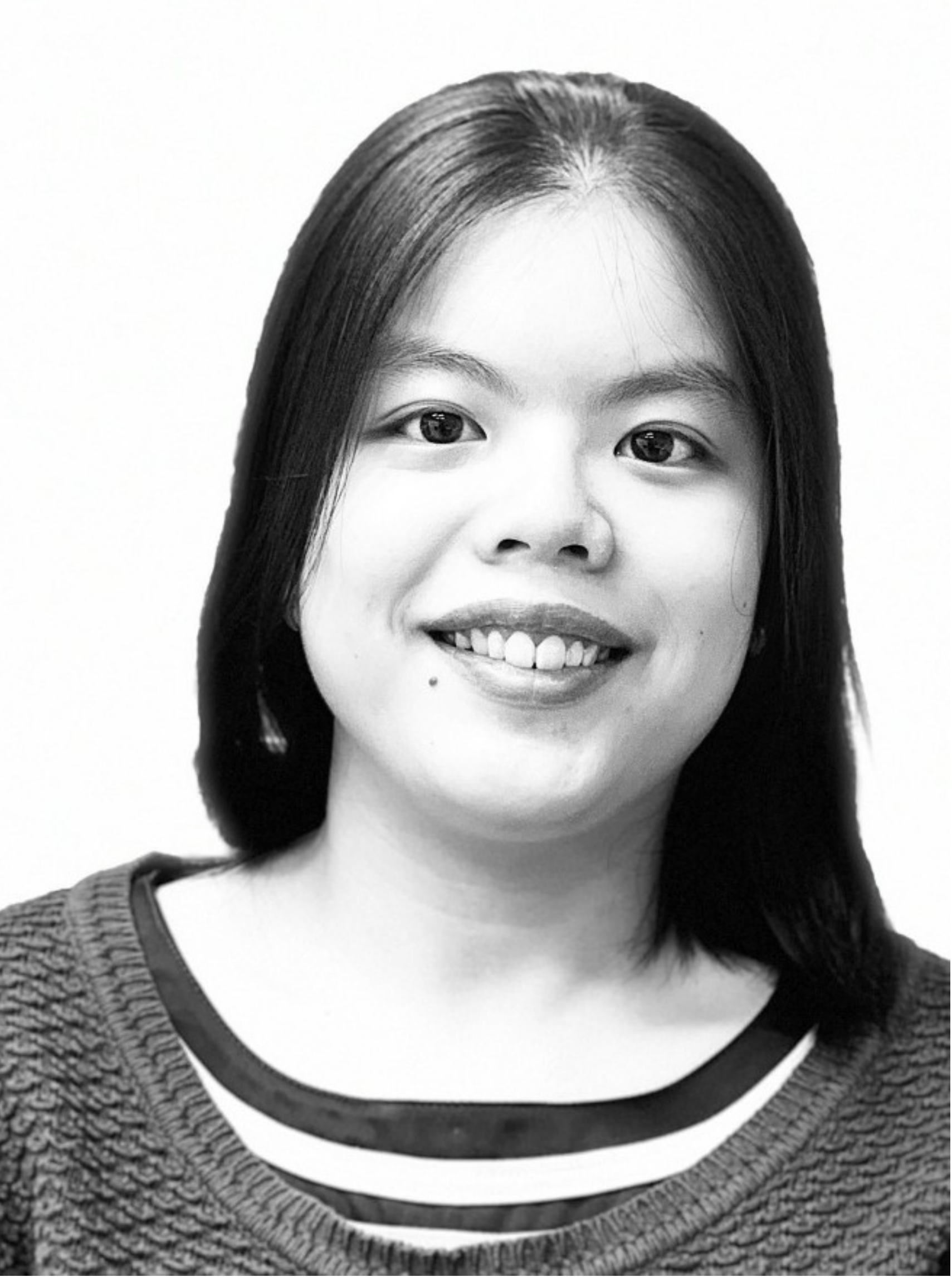}}]{Zann Koh} received the B.Eng degree in Engineering and Product Development from the Singapore University of Technology and Design, Singapore, in 2017. She is currently pursuing the Ph.D. degree with the Singapore University of Technology and Design, Singapore, under Dr. Chau Yuen’s supervision. Her current research interests include big data analysis, data discovery, urban human mobility, and unsupervised machine learning.
\end{IEEEbiography}
\begin{IEEEbiography}[{\includegraphics[width=1in,height=1.25in,clip,keepaspectratio]{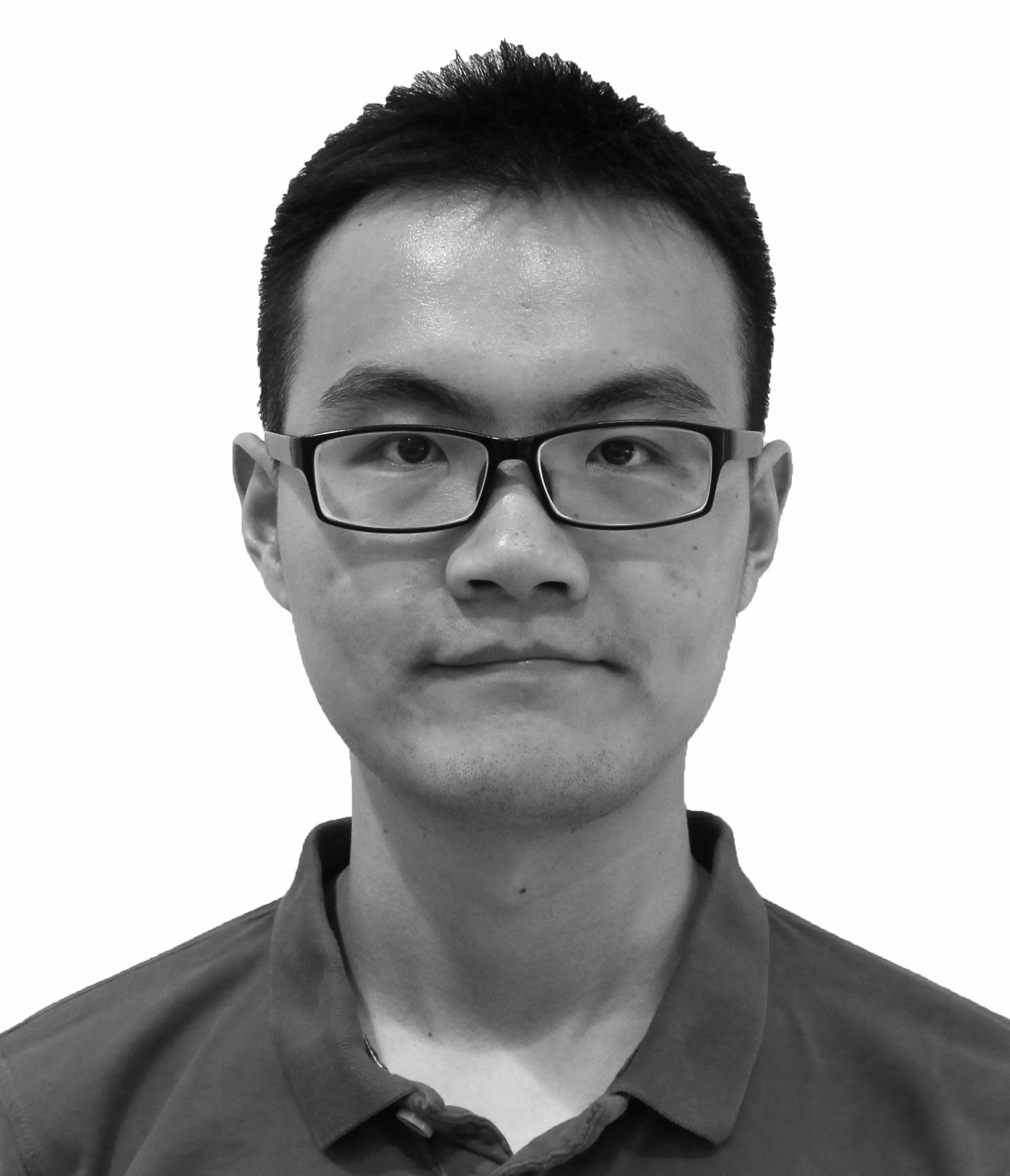}}]{Yuren Zhou} received the B.Eng. degree in Electrical Engineering from Harbin Institute of Technology,	Harbin, China in 2014, and the Ph.D. degree from Singapore University of Technology and Design, Singapore in 2019, with a focus on data mining and smart city applications. He is currently a postdoctoral research fellow at Singapore University of Technology and Design. His current research interests include big data analytics and its application in urban human mobility, building energy management, and Internet of Things.
\end{IEEEbiography}
\begin{IEEEbiography}[{\includegraphics[width=1in,height=1.25in,clip,keepaspectratio]{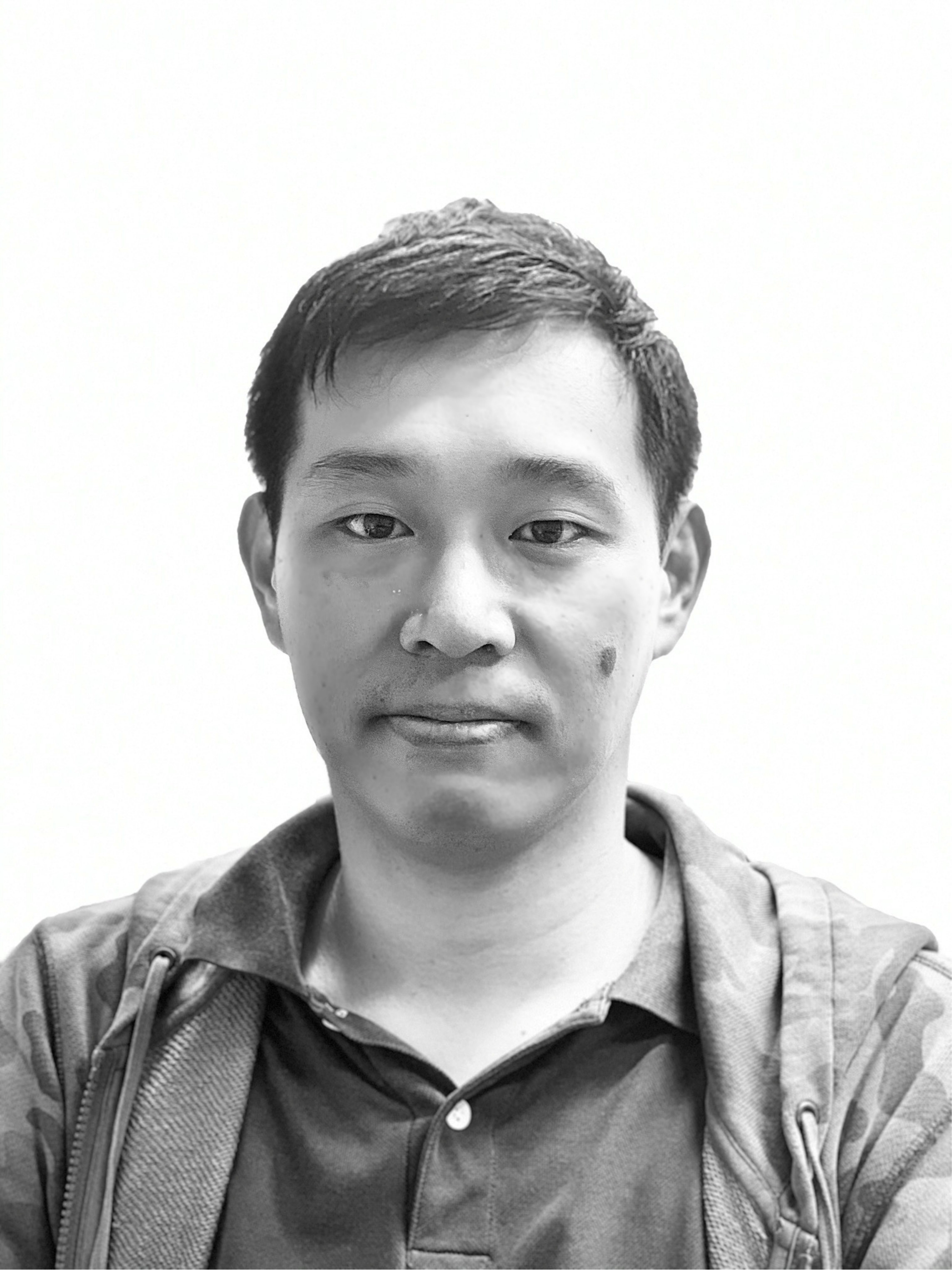}}]{Billy Pik Lik Lau} received the degree in computer science and M.Phil. degree in computer science from Curtin University, Perth, WA, Australia, in 2010 and 2014, respectively. He is currently a Ph.D. candidate with Dr. Chau Yuen at the Singapore University of Technology and Design, Singapore. He studied the cooperation rate between agents in multiagents systems during master studies. His current research interests include urban science, big data analysis, data knowledge discovery, Internet of Things, and unsupervised machine learning.
\end{IEEEbiography}
\begin{IEEEbiography}[{\includegraphics[width=1in,height=1.25in,clip,keepaspectratio]{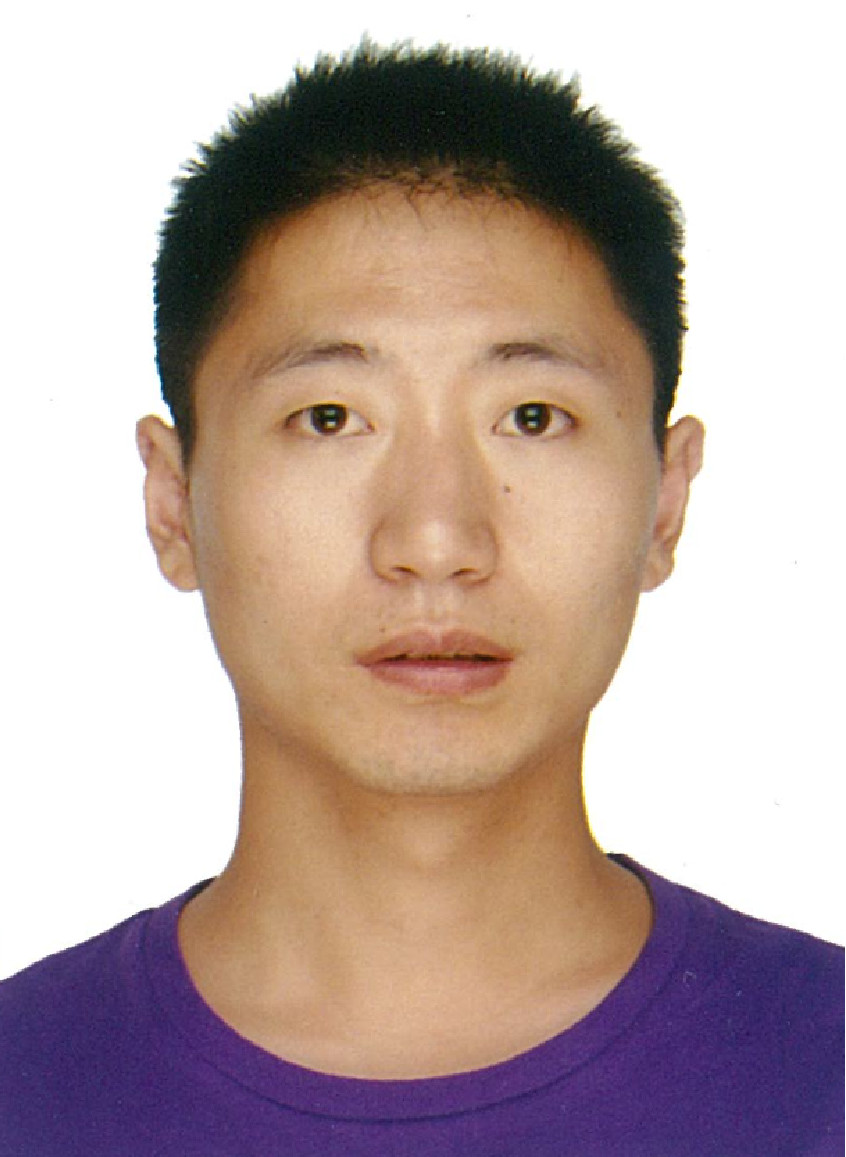}}]{Ran Liu} Ran Liu received the B.S. degree from the Southwest University of Science and Technology, China, in 2007, and the Ph.D. degree from the University of Tuebingen, Germany, in 2014, under the supervision of Prof. Andreas Zell. Since 2014, he has been a Research Fellow under the supervision of Prof. Chau Yuen with the MIT International Design Center, Singapore University of Technology and Design, Singapore. His research interests include robotics, indoor positioning, and SLAM.	
\end{IEEEbiography}
\begin{IEEEbiography}[{\includegraphics[width=1in,height=1.25in,clip,keepaspectratio]{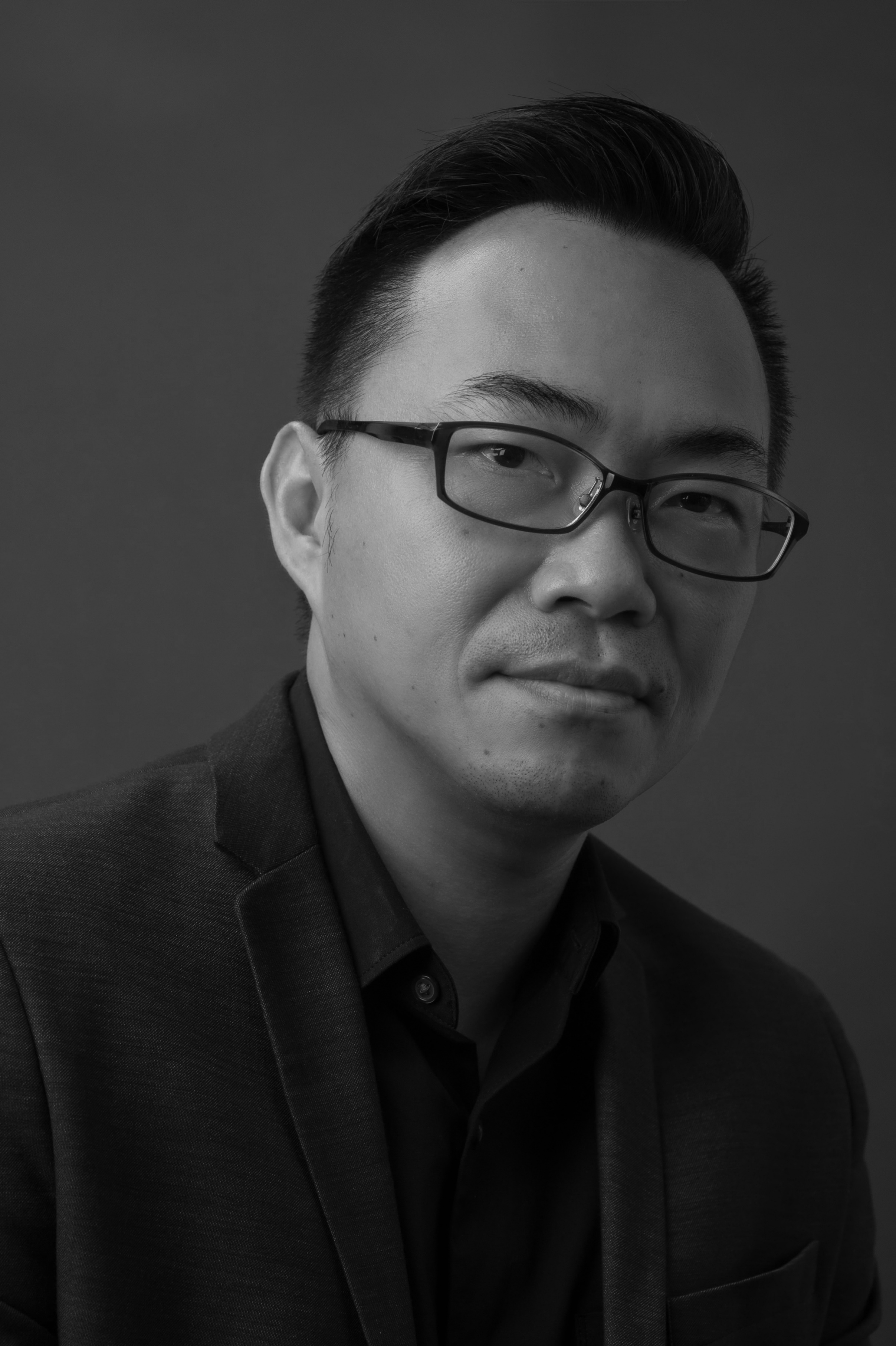}}]{Keng Hua Chong} is Associate Professor of Architecture and Sustainable Design at the Singapore University of Technology and Design (SUTD), where he directs the Social Urban Research Groupe (SURGe) and co-leads the Opportunity Lab (O-Lab). His research on social architecture particularly in the areas of ageing population, liveable place and data-driven collaborative design has led to several key publications and projects, including Creative Ageing Cities, Second Beginnings, and the New Urban Kampung Research Programme.
\end{IEEEbiography}
\begin{IEEEbiography}[{\includegraphics[width=1in,height=1.25in,clip,keepaspectratio]{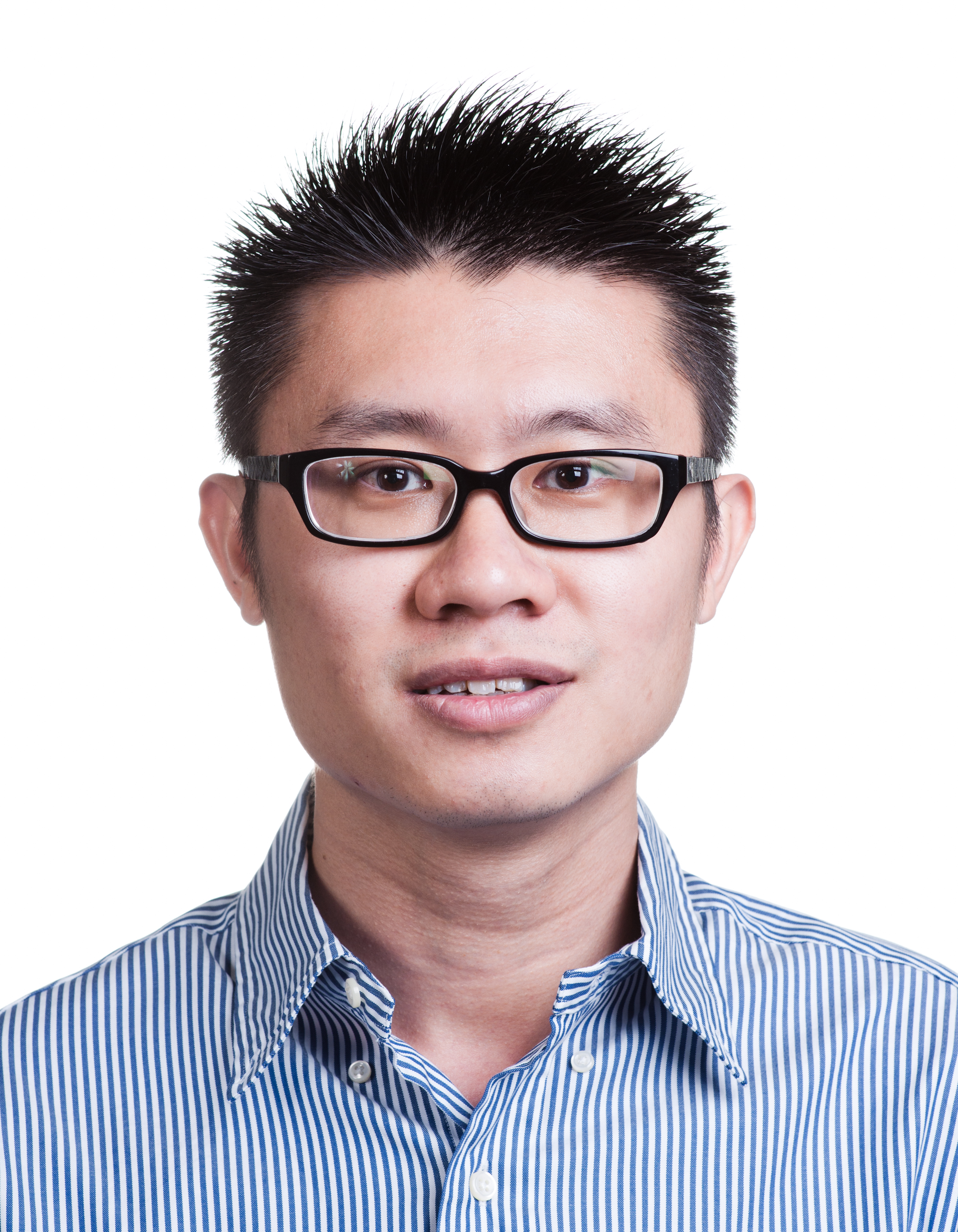}}]{Chau Yuen} is currently an Associate Professor at Singapore University of Technology and Design. He received the B.Eng. and Ph.D. degrees from Nanyang Technological University, Singapore, in 2000 and 2004, respectively. He was a Postdoctoral Fellow at Lucent Technologies Bell Labs, Murray Hill, NJ, USA, in 2005. He was a Visiting Assistant Professor at The Hong Kong Polytechnic University in 2008. From 2006 to 2010, he was a Senior Research Engineer at the Institute for Infocomm Research (I2R, Singapore), where he was involved in an industrial project on developing an 802.11n Wireless LAN system, and participated actively in 3Gpp Long Term Evolution (LTE) and LTE-Advanced (LTE-A) Standardization. He has been with the Singapore University of Technology and Design since 2010. He is a recipient of the Lee Kuan Yew Gold Medal, the Institution of Electrical Engineers Book Prize, the Institute of Engineering of Singapore Gold Medal, the Merck Sharp and Dohme Gold Medal, and twice the recipient of the Hewlett Packard Prize. He received the IEEE Asia-Pacific Outstanding Young Researcher Award in 2012. He serves as an Editor for the IEEE Transaction on Communications and the IEEE Transactions on Vehicular Technology and was awarded the Top Associate Editor from 2009 to 2015.
\end{IEEEbiography}

\EOD
	
\end{document}